\documentclass[11pt]{article}

\usepackage[
  shownumpages,               
  bgcolor={255,253,246},      
  braincolor={157,225,252},    
  linkcolor=teal,
  citecolor={144,199,225},  
  urlcolor=blue,            
  citingstyle=authoryear,     
  bibliostyle=plainnat,       
  bibfile=references          
]{brainlab}

\usepackage{mathtools}      

\usepackage{booktabs}       
\usepackage{multirow}
\usepackage{wrapfig}   
\usepackage{graphicx}  

\setbrainmeta{
  title={Scalable Knowledge Editing for Mixture-of-Experts LLMs via Tensor-Structured Updates},
  authors={
    Roman Maksimov\textsuperscript{1,2}, Vladimir Aletov\textsuperscript{1,2}, Dmitry Bylinkin\textsuperscript{1,2}, Daniil Medyakov\textsuperscript{1,2}, Vladimir Solodkin\textsuperscript{1,2}, Aleksandr Beznosikov\textsuperscript{1,2,3}
  },
  affiliations={
    \textsuperscript{1}Moscow Independent Research Institute of Artificial Intelligence \\
    \textsuperscript{2}Basic Research of Artificial Intelligence Laboratory \\
    \textsuperscript{3}Innopolis University
  },
  abstract={
    Knowledge editing (KE) provides a lightweight alternative to repeated fine-tuning of LLMs. However, most existing KE methods target dense feed-forward layers, while modern LLMs increasingly adopt Mixture-of-Experts (MoE) architectures for their superior memory footprint and inference efficiency. This mismatch leaves a growing class of production models without principled editing tools. We propose a MEMIT-like framework for knowledge editing in MoE-based LLMs. Our method exploits the tensor structure of MoE layers to formulate the editing objective faithfully at the per-expert level, and applies the Woodbury matrix identity to avoid materializing or inverting the full stacked matrix of expert weights. The resulting update reduces to inversions of fixed low-rank matrices and requires no additional backward passes. Empirically, our approach matches the editing quality of strong baselines on the main KE metrics while accelerating the editing procedure by up to 6x, owing to the batched MEMIT-style formulation and the low-dimensional inversions enabled by the Woodbury identity. These results show that closed-form, parameter-modifying KE can be extended efficiently beyond dense layers, opening a path toward scalable knowledge editing in modern sparse LLM architectures.
  },
}

\newcommand{\R}{\mathbb{R}}
\newcommand{\err}[1]{{\tiny $\pm$#1}}
\DeclareMathOperator*{\argmin}{arg\, min}

\begin{document}

\begin{mainpart}

\section{Introduction}
\label{sec:introduction}

Large language models (LLMs) have become foundational tools in a wide range of natural language processing applications. However, the factual knowledge encoded in their parameters is inherently static once training concludes, while the world in which they are deployed is not. Maintaining factual accuracy through repeated fine-tuning is prohibitively expensive at the scale of modern LLMs. This holds even for parameter-efficient methods, which still require substantial compute and curated training data~\citep{mitchell2021fast} and which risk catastrophic forgetting of previously acquired knowledge when applied repeatedly~\citep{luo2025empirical}. \\

\textbf{Knowledge editing} (KE)~\citep{wang2024knowledge} has emerged as a principled response to this challenge. Rather than retraining the model, KE methods aim to inject or overwrite individual factual associations through targeted modifications of model parameters or auxiliary modules. A particularly successful line of work, represented by ROME~\citep{meng2022locating} and its batched extension MEMIT~\citep{meng2022mass}, formulates editing as a closed-form least-squares update on the down-projection matrix of a feed-forward (FFN) layer, treated as a linear associative memory~\citep{anderson1972simple, geva2021transformer, sun2024head}. These methods are attractive because they are fast, they do not require gradient-based optimization of the edited weights, and admit a clean theoretical interpretation. \\

Following MEMIT~\citep{meng2022mass}, the field of knowledge editing has developed along several complementary directions. The most relevant to our work is the line that refines the closed-form \emph{locate-and-edit} paradigm: PMET~\citep{li2024pmet} jointly optimizes the hidden states of both MHSA and FFN sublayers but uses only the FFN component for the weight update, yielding more precise edits, while EMMET~\cite{gupta2024unified}  unifies ROME and MEMIT under a common preservation--memorization objective and proposes a batched equality-constrained variant. Within the same line, AlphaEdit~\cite{fang2024alphaedit} addresses the degradation of locate-and-edit methods under long sequences of edits by projecting each parameter perturbation onto the null-space of the preserved knowledge, thereby preventing interference between successive updates and substantially improving stability in the sequential-editing regime. Beyond this paradigm, complementary directions include tackling the reversal curse via bidirectional editing objectives~\citep{ma2023untying}, memory-based approaches that keep the base weights untouched and attach an external codebook or side memory~\cite{hartvigsen2023aging,wang2024wise}, and regularization-based methods that preserve the model's general abilities under many edits~\cite{gu2024model}. \\

A critical limitation is that essentially all closed-form parameter-editing methods have been developed and evaluated on \textit{dense} transformer architectures. Modern leading LLMs increasingly depart from this regime: state-of-the-art systems such as GPT-OSS~\cite{openai2025gptoss120bgptoss20bmodel}, DeepSeek-v4~\cite{deepseekai2026deepseekv4}, and Qwen$3.6$~\cite{qwen36_35b_a3b} adopt \textbf{Mixture-of-Experts} (MoE) designs~\citep{shazeer2017outrageously,molodtsov2026himoe}, in which the dense FFN is replaced by a router and a pool of experts, only a small subset of which is activated per token. MoE architectures dominate the current frontier for a clear reason: they decouple parameter count from per-token compute. This enables dramatically larger capacity at fixed inference cost. As a result, in modern MoE-based LLMs, the parameters that previously stored factual associations in dense FFNs are now distributed across a pool of experts gated by a router~\citep{zhao2024factorllm}. This mismatch raises a concrete and underexplored question:

\begin{center}
\emph{How can closed-form, parameter-modifying knowledge editing be performed faithfully and efficiently in MoE-based LLMs?}
\end{center}

To the best of our knowledge, the only method developed specifically for this setting is MoE-Edit~\citep{gu2026moeedit}. It reformulates the KE objective in a way that faithfully accounts for the MoE structure, and solves the resulting problem by applying MEMIT-like updates. Within each layer, an approximate block coordinate descent (BCD) procedure iterates over experts one at a time, computing per-expert updates that are subsequently projected onto the null-space of the preserved knowledge, following the AlphaEdit~\cite{fang2024alphaedit} formulation. However, this design imposes a twofold sequential bottleneck: \textit{(i)} activation collection must be repeated layer by layer for every batch of edits, while within each layer \textit{(ii)} the BCD solver iterates sequentially over experts. As the number of edits and the number of experts grow, both factors compound and make the procedure increasingly expensive, motivating a closed-form alternative. 

\paragraph{Contributions.} In this work, we propose a novel framework tailored to MoE architectures. Our contributions are as follows:

\begin{itemize}
    \item We formulate the knowledge editing objective for MoE layers in a way that respects the tensor structure of stacked expert weights, avoiding both single-expert and naively concatenated dense surrogates. Exploiting this structure, we recast the problem into a form where the Woodbury identity applies efficiently, yielding a closed-form update that never materializes or inverts the full stacked expert matrix and requires only inversions of size fixed by the number of edits.

    \item The resulting update rule of \textbf{MoE Tucker Editor} (\texttt{MoTE}) is fully closed-form and \textbf{backward-pass-free}: like MEMIT in the dense case, it requires no gradient computation through the edited layer and inherits the batched-edit property that makes MEMIT practical at scale.

    \item Empirically, \texttt{MoTE}  achieves editing quality comparable to strong baselines on efficacy, generalization, and specificity, while accelerating the editing procedure by \textbf{up to $\mathbf{6\times}$}. This speedup is achieved by collecting activations only once at the last layer and finding closed form solution only once for that last layer.
\end{itemize}

The remainder of the paper is organized as follows. Section~\ref{sec:preliminaries} introduces background on MEMIT and the structure of MoE FFN layers. Section~\ref{sec:method} presents our method. Section~\ref{sec:experiments} reports experimental results on standard KE benchmarks, and Section~\ref{sec:conclusion} concludes with a discussion of limitations and future directions.

\section{Preliminaries}
\label{sec:preliminaries}
\subsection{Algebraic prerequisites}
\label{subsec:linal}

\paragraph{The Woodbury identity.}
Let $A \in \R^{n \times n}$ be invertible, $U \in \R^{n \times T}$, $V \in \R^{T \times n}$, and $C \in \R^{T \times T}$ invertible. The Woodbury identity~\citep{hager1989updating, duncan1944lxxviii} states
\begin{equation*}
(A + U C V)^{-1}
\;=\; A^{-1} - A^{-1} U \bigl( C^{-1} + V A^{-1} U \bigr)^{-1} V A^{-1}.
\end{equation*}
Specializing $A = \lambda I_n$, $C = I_T$, $U = \Psi$, and $V = \Psi^{\top}$ yields the \emph{push-through corollary}:
\begin{equation}
\label{eq:push-through}
\Psi^{\top} \bigl( \Psi \Psi^{\top} + \lambda I_n \bigr)^{-1}
\;=\;
\bigl( \Psi^{\top} \Psi + \lambda I_T \bigr)^{-1} \Psi^{\top},
\qquad \forall\, \lambda > 0,\; \Psi \in \R^{n \times T},
\end{equation}
which trades an $n \times n$ inversion for a $T \times T$ inversion at the cost of two matrix multiplications.

\paragraph{Tucker Decomposition.}
Let $\mathcal{X} \in \R^{I_1 \times I_2 \times \cdots \times I_N}$ be an $N$-th order tensor. A \emph{Tucker decomposition}~\citep{tucker1963implications} factors $\mathcal{X}$ into a smaller \emph{core tensor} $\mathcal{G} \in \R^{R_1 \times R_2 \times \cdots \times R_N}$ multiplied along each mode by a \emph{factor matrix} $U^{(n)} \in \R^{I_n \times R_n}$:
\begin{equation}
\label{eq:tucker}
\mathcal{X} \;\approx\; \mathcal{G} \times_1 U^{(1)} \times_2 U^{(2)} \times_3 \cdots \times_N U^{(N)},
\end{equation}
where $\times_n$ denotes the mode-$n$ product.  The tuple $(R_1, \dots, R_N)$ is the \emph{multilinear rank} of the decomposition. Choosing $R_n < I_n$ enforces a low-rank structure along mode $n$. The factors $U^{(n)}$ are typically required to have orthonormal columns, in which case~\eqref{eq:tucker} is known as the \emph{higher-order SVD} (HOSVD)~\citep{de2000multilinear}, computed by truncating the SVD of each unfolding $X_{(n)}$.

\paragraph{Tucker decompositions in deep learning.}
Tucker decomposition has long served as a structural prior on neural-network parameters. The natural modes of trained neural-network weight tensors -- channels, layers, heads, or experts -- exhibit a multilinear rank far below their nominal dimension~\citep{gu2025tensorllm, peshekhonov2024training, li2026lestd, yuebintd}.

\subsection{Dense FFN as a Linear Associative Memory}
\label{subsec:dense-ffn}

A standard Transformer block alternates a multi-head self-attention (MHSA) sublayer with a position-wise feed-forward network (FFN). In modern gated variants, the FFN consists of three linear maps: an \emph{up-projection} $W_{\mathrm{up}} \in \R^{d_{\mathrm{hidden}} \times d_{\mathrm{model}}}$, a \emph{gate projection} $W_{\mathrm{gate}} \in \R^{d_{\mathrm{hidden}} \times d_{\mathrm{model}}}$, and a \emph{down-projection} $W_{\mathrm{down}} \in \R^{d_{\mathrm{model}} \times d_{\mathrm{hidden}}}$.

Following~\citep{geva2021transformer, meng2022locating}, the down-projection admits a natural interpretation as a \emph{linear associative memory}: each of its $d_{\mathrm{hidden}}$ columns acts as a value vector associated with a learned key direction in the gated activation space. Concretely, given a hidden state $u \in \R^{d_{\mathrm{model}}}$, the FFN first builds a key
\begin{equation*}
k = \sigma\!\left( W_{\mathrm{gate}}\, u \right) \odot \left( W_{\mathrm{up}}\, u \right) \;\in\; \R^{d_{\mathrm{hidden}}},
\end{equation*}
where $\sigma(\cdot)$ is a non-linearity and $\odot$ denotes the Hadamard product. The FFN output is then a linear combination of the columns of $W_{\mathrm{down}}$ weighted by $k$:
\begin{equation*}
v = W_{\mathrm{down}}\, k \;\in\; \R^{d_{\mathrm{model}}}.
\end{equation*}

\paragraph{Knowledge editing as a regularized least-squares problem.}
Knowledge editing seeks to install a new association into this memory. For a target fact $(s, r, o_{\mathrm{new}})$ -- a subject $s$, a relation $r$, and a desired new object $o_{\mathrm{new}}$ -- the goal is to modify the model so that the key $k_{(s,r)}$ corresponding to the prompt encoding $(s, r)$ is mapped to a new target vector $z \in \R^{d_{\mathrm{model}}}$ that steers downstream computation toward $o_{\mathrm{new}}$. Two competing requirements must be balanced. The update should be \emph{effective} on the edited fact and its paraphrases (efficacy and generalization), and \emph{local}, i.e.\ leave the model's predictions on unrelated inputs essentially unchanged (specificity).

In the dense case, this trade-off is naturally expressed as a regularized least-squares problem over an additive perturbation $\Delta \in \R^{d_{\mathrm{model}} \times d_{\mathrm{hidden}}}$:
\begin{equation}
\label{eq:dense-edit}
\Delta^{\star}
= \argmin_{\Delta}\;
\underbrace{\bigl\| (W_{\mathrm{down}} + \Delta)\, K_1 - V_1 \bigr\|_F^2}_{\text{memorization}}
\;+\;
\underbrace{\bigl\| \Delta\, K_0 \bigr\|_F^2}_{\text{preservation}}
\;+\;
\lambda\, \| \Delta \|_F^2,
\end{equation}
where $K_1 \in \R^{d_{\mathrm{hidden}} \times T}$ stacks the keys of the $T$ edited facts column-wise, $V_1 \in \R^{d_{\mathrm{model}} \times T}$ contains the corresponding target values, and $K_0 \in \R^{d_{\mathrm{hidden}} \times M}$ is a preservation key set sampled from the model's empirical activation distribution on a large generic corpus. The Frobenius regularizer $\lambda > 0$ controls the magnitude of the update. Problem~\eqref{eq:dense-edit} admits a closed-form solution involving only a single $d_{\mathrm{hidden}} \times d_{\mathrm{hidden}}$ inversion, which underlies the ROME~\citep{meng2022locating} and MEMIT~\citep{meng2022mass} families of methods.

\subsection{MoE Layer as a Mixture of Expert Associative Memories}
\label{subsec:moe-layer}

Modern frontier LLMs replace the dense FFN with a \emph{Mixture-of-Experts} (MoE) sublayer~\citep{shazeer2017outrageously, lepikhin2020gshard, du2022glam}. The motivation is well known: MoE decouples a model's parameter count from its per-token compute. A pool of $E$ specialized FFN-shaped experts collectively stores far more parameters than a dense FFN of the same compute budget, while only $K \ll E$ of them are activated per token via a sparse routing mechanism.

\paragraph{Router.}
At the core of the layer is a \emph{router}, a lightweight learned function that determines which experts process each token. Given an input hidden state $x \in \R^{d_{\mathrm{model}}}$, the router produces affinity logits with respect to $E$ learned expert embeddings $e_1, \dots, e_E \in \R^{d_{\mathrm{model}}}$:
\begin{equation*}
s_j = x^{\top} e_j, \qquad j = 1, \dots, E.
\end{equation*}
A top-$K$ operator selects the indices of the $K$ largest logits,
\begin{equation*}
S = \mathrm{TopK}\bigl( \{s_1, \dots, s_E\} \bigr) \subset \{1, \dots, E\}, \qquad |S| = K,
\end{equation*}
and the corresponding logits are renormalized into a sparse gating distribution over the selected experts:
\begin{equation}
\label{eq:gating}
g_j =
\begin{cases}
\dfrac{\exp(s_j)}{\sum_{\ell \in S} \exp(s_{\ell})}, & j \in S, \\[8pt]
0, & j \notin S.
\end{cases}
\end{equation}
By construction, $\sum_{j} g_j = 1$ and the gating vector $\mathbf{g} = (g_1, \dots, g_E)$ is $K$-sparse.

\paragraph{Experts.}
Each expert $j \in \{1, \dots, E\}$ implements a self-contained gated FFN with its own parameters \\ $\bigl( W^{(j)}_{\mathrm{gate}},\; W^{(j)}_{\mathrm{up}},\; W^{(j)}_{\mathrm{down}} \bigr)$. By analogy with the dense case, expert $j$ first computes its own key
\begin{equation*}
k_j = \sigma\!\left( W^{(j)}_{\mathrm{gate}}\, x \right) \odot \left( W^{(j)}_{\mathrm{up}}\, x \right) \;\in\; \R^{d_{\mathrm{hidden}}},
\end{equation*}
and then produces its contribution by reading out from its own associative memory:
\begin{equation*}
v_j(x) = W_j\, k_j, \qquad W_j \coloneqq W^{(j)}_{\mathrm{down}} \in \R^{d_{\mathrm{model}} \times d_{\mathrm{hidden}}}.
\end{equation*}
Crucially, both the keys $k_j$ \emph{and} the values $v_j $ are expert-specific: different experts construct different gated activations from the same input $u$ and project them through different down-projections.

\paragraph{Mixture output.}
The output of the MoE layer is the router-weighted sum of the contributions of the active experts
\begin{equation}
\label{eq:moe-output}
v = \sum_{j=1}^{E} g_j\, v_j = \sum_{j \in S} g_j\, W_j\, k_j,
\end{equation}
where the second equality follows from the sparsity of $\mathbf{g}$ in~\eqref{eq:gating}. Equation~\eqref{eq:moe-output} makes explicit the structural property that distinguishes MoE editing from the dense case: a single MoE layer is no longer one associative memory $W_{\mathrm{down}}$, but a \emph{mixture of $E$ associative memories} $\{W_j\}_{j=1}^{E}$, dynamically combined per token by the router.

\paragraph{Closed-form MoE editing objective.}
\citep{gu2026moeedit} recently extended the dense formulation~\eqref{eq:dense-edit} to MoE layers. Treating the per-expert down-projections $\{W_j\}_{j=1}^E$ jointly as the optimization variable, they cast knowledge editing as the following block-structured regularized problem:
\begin{equation}
\label{eq:moe-edit-obj}
\bigl\{ \Delta_j^{\star} \bigr\}_{j=1}^{E}
\;=\; \argmin_{\{\Delta_j\}}\;
\sum_{f \in \mathcal{F}}
\!\left\| \sum_{j=1}^{E} g_{f,j}\, (W_j + \Delta_j)\, k_{f,j} - v_f \right\|^2
\!+ \sum_{f \in \mathcal{P}}
\!\left\| \sum_{j=1}^{E} g_{f,j}\, \Delta_j\, k_{f,j} \right\|^2
\!+ \lambda \sum_{j=1}^{E} \| \Delta_j \|^2,
\end{equation}
where $\mathcal{F}$ is the set of facts to edit, $\mathcal{P}$ the preservation set, and $g_{f,j}$, $k_{f,j}$ the router weight and expert-specific key induced by prompt $f$ on expert $j$. Stacking the per-expert updates into the vector $\theta = \mathrm{vec}\!\bigl( [\Delta_1\, \cdots\, \Delta_E] \bigr)$ and defining the per-fact design vector
\begin{equation*}
\psi_f = \bigl[ g_{f,1}\, k_{f,1}^{\top} \;\; \cdots \;\; g_{f,E}\, k_{f,E}^{\top} \bigr]^{\top} \in \R^{E\, d_{\mathrm{hidden}}},
\qquad
r_f = v_f - \sum_{j=1}^{E} g_{f,j}\, W_j\, k_{f,j},
\end{equation*}
problem~\eqref{eq:moe-edit-obj} admits the closed-form solution
\begin{align}
\theta^{\star}
&= \left( \sum_{f \in \mathcal{F}} (\psi_f \psi_f^{\top}) \otimes I_{d_{\mathrm{model}}} \;+\; \lambda I \right)^{-1}
\sum_{f \in \mathcal{F}} (\psi_f \otimes I_{d_{\mathrm{model}}}) \, r_f, \\
\Delta^{\star} &= \texttt{unvec}(\theta^{\star}),
\end{align}
where $\otimes$ denotes the Kronecker product. Although elegant, this global solution requires inverting a system of size $(E\, d_{\mathrm{hidden}}) \times (E\, d_{\mathrm{hidden}})$, which is computationally prohibitive at realistic scales. This limitation motivates the block-coordinate treatment developed in the MoE-Edit work.



\section{Method}
\label{sec:method}

\begin{wrapfigure}[13]{r}{0.4\textwidth}
    \vspace{-5mm}
  \centering
  \includegraphics[width=\linewidth]{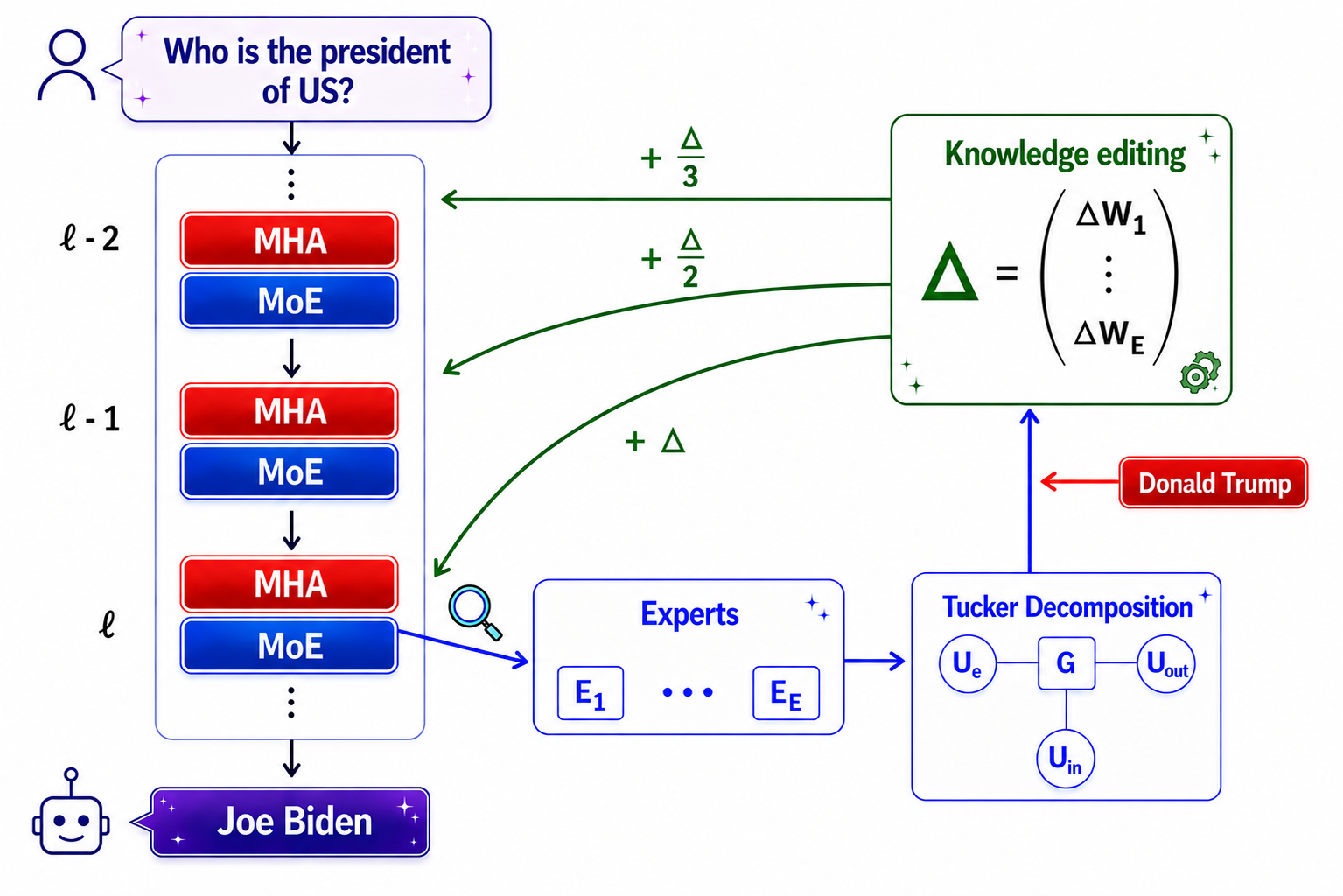}
  \caption{MoTE editing pipeline}
  \label{fig:example}
\end{wrapfigure}
Section~\ref{sec:preliminaries} ended with the closed-form MoE editing objective~\eqref{eq:moe-edit-obj}, whose unique global minimizer requires solving a linear system of size $(E\,d_{\mathrm{hidden}})\times(E\,d_{\mathrm{hidden}})$ -- intractable at modern MoE scales. We construct \texttt{MoTE} by composing three ingredients. \textbf{(i)} A \emph{Woodbury reduction} that shrinks the per-layer inversion to size $T\times T$, where $T$ is the number of edits in a batch, independent of $E$ and $d_{\mathrm{hidden}}$. \textbf{(ii)} A \emph{MEMIT-style updates schedule} that replaces the per-layer solve with spreading deltas across layers. \textbf{(iii)} A \emph{Tucker reparameterization} of the update tensor $\Delta\mathcal{W}\in\R^{E\times d_{\mathrm{model}}\times d_{\mathrm{hidden}}}$ that respects the underlying geometric structure of the MoE down-projection. As we show in Section~\ref{subsec:woodbury-memit}, the first ingredient in cope with MEMIT residual schedule can yield a direct MoE adaptation that is computationally tractable but empirically uncompetitive; ingredients~(ii) and ~(iii), developed in Section~\ref{subsec:tucker-memit}, supply the missing structural prior and provide significant speed-up.

\subsection{Woodbury Reduction and Multi-layer Spread}
\label{subsec:woodbury-memit}
The closed-form solution of~\eqref{eq:moe-edit-obj} requires inverting a system of size $(E\,d_{\mathrm{hidden}})\times(E\,d_{\mathrm{hidden}})$, with time and memory costs of $\mathcal{O}((E\,d_{\mathrm{hidden}})^3)$ and $\mathcal{O}((E\,d_{\mathrm{hidden}})^2)$ respectively. However, the Gram matrix driving the inversion is structurally low-rank: it is a sum of $T=|\mathcal{F}|$ rank-one terms $\psi_f \psi_f^{\top}$, where $T$ -- the number of edits processed in a batch -- is several orders of magnitude smaller than $E\,d_{\mathrm{hidden}}$. This is precisely the regime in which the Woodbury identity~\eqref{eq:push-through} converts the inversion into a much smaller one.

Stacking the per-fact design vectors and target residuals as
\[
\Psi = \bigl[ \psi_1\;\cdots\;\psi_T \bigr]\in\R^{E\,d_{\mathrm{hidden}}\times T},
\qquad
R = \bigl[ r_1\;\cdots\;r_T \bigr]\in\R^{d_{\mathrm{model}}\times T},
\]
and unvectorizing $\theta^{\star}$ into the matrix $\hat{\Delta}^{\star}=[\Delta_1^{\star}\;\cdots\;\Delta_E^{\star}]\in\R^{d_{\mathrm{model}}\times E\,d_{\mathrm{hidden}}}$, the global minimizer of~\eqref{eq:moe-edit-obj} reads
\begin{equation*}
\hat{\Delta}^{\star} \;=\; R\,\Psi^{\top}\bigl(\Psi\Psi^{\top}+\lambda I_{E\,d_{\mathrm{hidden}}}\bigr)^{-1}.
\end{equation*}
Substituting the push-through corollary~\eqref{eq:push-through} into the rightmost factor yields the equivalent but computationally far cheaper expression
\begin{equation}
\label{eq:moe-woodbury}
\boxed{\;
\hat{\Delta}^{\star} \;=\; R\,\bigl(\Psi^{\top}\Psi+\lambda I_T\bigr)^{-1}\Psi^{\top}.
\;}
\end{equation}
The only matrix to invert is now of size $T\times T$, independent of both the number of experts $E$ and the expert hidden dimension $d_{\mathrm{hidden}}$. Concretely,~\eqref{eq:moe-woodbury} requires forming the kernel matrix $\Psi^{\top}\Psi\in\R^{T\times T}$ in $\mathcal{O}(T^2\,E\,d_{\mathrm{hidden}})$ time, factorizing a $T\times T$ system in $\mathcal{O}(T^3)$ time, and assembling $\hat{\Delta}^{\star}$ via two dense products in $\mathcal{O}(T\,d_{\mathrm{model}}\,E\,d_{\mathrm{hidden}})$ time.

\paragraph{Plugging Woodbury into a MEMIT-style multi-layer editor.}
Equation~\eqref{eq:moe-woodbury} solves the editing problem at a single critical layer, but locate-then-edit methods derive most of their leverage by spreading an edit across several layers. The naive and natural approach here is to adapt the standard MEMIT~\citep{meng2022mass} schedule. Having solved~\eqref{eq:moe-edit-obj} at the topmost critical layer $L$ via~\eqref{eq:moe-woodbury}, we obtain per-fact target residuals
\begin{equation*}
r_{f}^{L} \;=\; v_f - \sum_{j=1}^{E} g_{f,j}^{L}\,W_{j}^{L}\,k_{f,j}^{L},
\end{equation*}
each of which already aggregates the contributions of all $E$ experts at layer $L$. We then propagate the edit downward by spreading these residuals across the preceding critical layers $\ell\in\{L_0,\dots,L\}$ with a coefficient that decays with the distance from $L$:
\begin{equation}
\label{eq:memit-spread}
r_{f}^{\ell} \;=\; \frac{r_{f}^{L}}{L-\ell+1},
\qquad
\hat{\Delta}^{\ell,\star} \;=\; R^{\ell}\bigl(\Psi^{L}\bigr)^{\!\top}\!\Bigl(\Psi^{L}(\Psi^{L})^{\!\top}+\lambda I\Bigr)^{-1},
\end{equation}
where $R^{\ell}=[r_{f}^{\ell}]_{f\in\mathcal{F}}\in\R^{d_{\mathrm{model}}\times T}$ stacks the per-layer residuals.
 Direct adaptation of MEMIT would require substituting $\Psi^{L}$ in \eqref{eq:memit-spread} with $\Psi^{\ell}\in\R^{E\,d_{\mathrm{hidden}}\times T}$ which stacks the per-fact design vectors $\psi_{f}^{\ell}=[g_{f,1}^{\ell}\,k_{f,1}^{\ell\,\top}\;\cdots\;g_{f,E}^{\ell}\,k_{f,E}^{\ell\,\top}]^{\!\top}$ -- the same construction as in~\eqref{eq:moe-edit-obj}, but with the router weights $g_{f,j}^{\ell}$ and per-expert keys $k_{f,j}^{\ell}$ recomputed at layer $\ell$, since both the routing distribution and the expert activations differ from layer to layer. Opposite to that, our approach allows to avoid additional computations by calculating the solution only once at the last layer and spreading it across previous layers with the same schedule.
 The denominator $L-\ell+1$ assigns the largest share to the layer closest to $L$ and gradually attenuates earlier layers, mirroring the original MEMIT schedule. Crucially, each per-layer solve in~\eqref{eq:memit-spread} has the form addressed by~\eqref{eq:moe-woodbury}: applying push-through gives the equivalent
\(
\hat{\Delta}^{\ell,\star} = R^{\ell}\bigl((\Psi^{L})^{\!\top}\Psi^{L}+\lambda I_T\bigr)^{-1}(\Psi^{L})^{\!\top},
\)
so the only matrices ever inverted across the entire multi-layer pipeline are of size $T\times T$. In this sense, the Woodbury reduction makes possible a direct MoE port of MEMIT -- and, more broadly, of any locate-then-edit method that relies on an analogous closed-form ridge solve -- computationally tractable out of the box.

\subsection{The Tensor-Structure Diagnosis}
\label{subsec:diagnosis}

\paragraph{An empirical surprise.}
Despite its apparent elegance, the direct combination of the Woodbury reduction~\eqref{eq:moe-woodbury} with the MEMIT-style multi-layer spread~\eqref{eq:memit-spread} fails to deliver competitive editing quality on MoE backbones. As we report in Section~\ref{sec:experiments}, this naive adaptation lags substantially behind dense-model expectations on the evaluation. This raises a natural question: 

\begin{center}
\emph{What aspect of the MoE architecture is this otherwise principled recipe failing to account for?}
\end{center}

We argue that the diagnosis is structural rather than algorithmic. By stacking the per-expert updates $\{\Delta_j\}_{j=1}^{E}$ into a flat matrix $\hat{\Delta}\in\R^{d_{\mathrm{model}}\times E\,d_{\mathrm{hidden}}}$, problem~\eqref{eq:moe-edit-obj} implicitly treats the $E$ experts as independent blocks of an otherwise unstructured linear map. The natural object underlying an MoE layer, however, is not a matrix but a third-order tensor: the per-expert down-projections $\{W_j\}_{j=1}^{E}$ assemble into
\begin{equation*}
\mathcal{W} \;\in\; \R^{E\times d_{\mathrm{model}}\times d_{\mathrm{hidden}}},
\end{equation*}
and the corresponding updates are naturally represented as tensors $\Delta\mathcal{W}$. Experts within the same layer are far from arbitrary: they share substantial structure inherited from joint pretraining. The flat block-diagonal solve of~\eqref{eq:memit-spread} discards this structure entirely, spending its regularization budget on update directions that lie \emph{outside} the low-multilinear-rank manifold along which experts actually vary; symmetrically, perturbations along that manifold are damped by the Frobenius regularizer at the same rate as genuinely unstructured noise. This observation motivates a tensor-aware reformulation, in which the unknown $\Delta\mathcal{W}$ is sought directly in factored form. A Tucker decomposition~\citep{tucker1963implications} is the canonical choice: it disentangles the three modes -- expert, model, and hidden -- through a small core tensor $\mathcal{G}$ and three mode-specific factor matrices, aligning the optimization variable with the multilinear geometry of the MoE layer.

\subsection{Tucker-Structured MoE Editor}
\label{subsec:tucker-memit}

We now combine the two ingredients on the table. The Woodbury reduction of Section~\ref{subsec:woodbury-memit} makes a per-layer MEMIT solve cheap in the unstructured case, while the diagnosis of Section~\ref{subsec:diagnosis} indicates that the unstructured solve is the wrong object: experts co-vary along a low-multilinear-rank manifold of $\mathcal{W}$. We constrain the editing update $\Delta\mathcal{W}$ to live on the Tucker manifold of $\mathcal{W}$ itself, and solve the resulting MEMIT objective in closed form via the Woodbury push-through.

Stack the per-expert updates into a tensor $\Delta\mathcal{W}\in\R^{E\times d_{\mathrm{model}}\times d_{\mathrm{hidden}}}$ with $\Delta\mathcal{W}[j,:,:]=\widehat{\Delta W}_j$, and constrain it to Tucker form
\begin{equation*}
\Delta\mathcal{W} \;=\; \mathcal{G}\times_1 U_e \times_2 U_{\mathrm{out}} \times_3 U_{\mathrm{in}},
\end{equation*}
with factors $U_e\in\R^{E\times r_e}$, $U_{\mathrm{out}}\in\R^{d_{\mathrm{model}}\times r_{\mathrm{out}}}$, $U_{\mathrm{in}}\in\R^{d_{\mathrm{hidden}}\times r_{\mathrm{in}}}$ and core $\mathcal{G}\in\R^{r_e\times r_{\mathrm{out}}\times r_{\mathrm{in}}}$, where $(r_e,r_{\mathrm{out}},r_{\mathrm{in}})\ll(E,d_{\mathrm{model}},d_{\mathrm{hidden}})$. The per-expert update reads
\begin{equation}
\label{eq:per-expert-update}
\widehat{\Delta W}_j \;=\; U_{\mathrm{out}}\Bigl(\sum_{a=1}^{r_e} U_e[j,a]\,G_a\Bigr) U_{\mathrm{in}}^{\!\top},
\qquad G_a \coloneqq \mathcal{G}[a,:,:].
\end{equation}
The factors $U_e, U_{\mathrm{out}}, U_{\mathrm{in}}$ are fixed, whereas the core $\mathcal{G}$ with $r_e r_{\mathrm{out}} r_{\mathrm{in}}$ free entries is optimized.

\paragraph{Factor extraction.}
We obtain $U_e, U_{\mathrm{out}}, U_{\mathrm{in}}$ once at initialization by HOSVD~\citep{de2000multilinear} on $\mathcal{W}$ itself, computed via the Gram trick: $U_n=\mathrm{TopEig}_{r_n}(\mathcal{W}_{(n)}\mathcal{W}_{(n)}^{\!\top})$ for $n\in\{e,\mathrm{out},\mathrm{in}\}$, optionally refined by a few HOOI~\citep{kolda2009tensor} sweeps. Following TD-MoE~\citep{yuebintd}, we additionally support multilinear whitening of $\mathcal{W}$ before HOSVD using mode-wise activation covariances $\Sigma_{\mathrm{out}},\Sigma_{\mathrm{in}}$, so that the factors are aligned with the data geometry rather than with the raw weight geometry; the recovered factors are subsequently re-coloured to the original space. We refer to the paper~\citep{yuebintd} for the full whitening recipe. Ablation on whitening types in Appendix \ref{appendix:whitening} shows that \textit{in} regime performs the best.

\paragraph{Compressed editing objective.}
Substituting~\eqref{eq:per-expert-update} into the per-layer MEMIT objective~\eqref{eq:memit-spread} collapses the $E\,d_{\mathrm{model}}\,d_{\mathrm{hidden}}-$ dimensional optimization onto the much smaller core $\mathcal{G}$. With null-space-projected keys $\widetilde{k}_{t,j}=P_j k_{t,j}$, define the \emph{compressed key} and the \emph{expert-mode aggregate}
\begin{equation*}
c_{t,j} \;=\; U_{\mathrm{in}}^{\!\top}\widetilde{k}_{t,j}\in\R^{r_{\mathrm{in}}},
\qquad
\phi_{t,a} \;=\; \sum_{j=1}^{E} g_{t,j}\,U_e[j,a]\,c_{t,j}\in\R^{r_{\mathrm{in}}},
\end{equation*}
and stack them as $\phi_t=[\phi_{t,1};\dots;\phi_{t,r_e}]\in\R^{r_e r_{\mathrm{in}}}$. Letting $G_{\mathrm{flat}}=[G_1\;\cdots\;G_{r_e}]\in\R^{r_{\mathrm{out}}\times r_e r_{\mathrm{in}}}$ denote the mode-1 unfolding of $\mathcal{G}$, a direct calculation gives
\begin{equation*}
U_{\mathrm{out}}^{\!\top}\sum_{j=1}^{E} g_{t,j}\,\widehat{\Delta W}_j\,\widetilde{k}_{t,j} \;=\; G_{\mathrm{flat}}\,\phi_t,
\end{equation*}
so that, after compressing the per-fact target residual analogously, $\widetilde{r}_t=U_{\mathrm{out}}^{\!\top}r_t\in\R^{r_{\mathrm{out}}}$, the projected MEMIT objective reduces to a ridge regression in the core variable:
\begin{equation}
\label{eq:tucker-objective}
\min_{G_{\mathrm{flat}}}\;\sum_{t=1}^{T}\bigl\|\widetilde{r}_t - G_{\mathrm{flat}}\phi_t\bigr\|_2^2 \;+\; \lambda \bigl\|G_{\mathrm{flat}}\bigr\|_F^2.
\end{equation}

Let $\Phi=[\phi_1,\dots,\phi_T]^{\!\top}\in\R^{T\times r_e r_{\mathrm{in}}}$ and $\widetilde{R}=[\widetilde{r}_1,\dots,\widetilde{r}_T]^{\!\top}\in\R^{T\times r_{\mathrm{out}}}$. The normal equations of~\eqref{eq:tucker-objective} read $G_{\mathrm{flat}}(\Phi^{\!\top}\Phi+\lambda I_{r_e r_{\mathrm{in}}})=\widetilde{R}^{\!\top}\Phi$, with unique minimizer
\begin{equation*}
G_{\mathrm{flat}}^{\star}
\;=\;\widetilde{R}^{\!\top}\Phi\,\bigl(\Phi^{\!\top}\Phi+\lambda I_{r_e r_{\mathrm{in}}}\bigr)^{-1}
\;=\;\widetilde{R}^{\!\top}\bigl(\Phi\Phi^{\!\top}+\lambda I_T\bigr)^{-1}\Phi,
\end{equation*}
where the second equality is the Woodbury push-through~\eqref{eq:push-through}. Either form yields the same $G_{\mathrm{flat}}^{\star}\in\R^{r_{\mathrm{out}}\times r_e r_{\mathrm{in}}}$, and we always solve whichever side has the smaller inversion: $T\times T$ in the typical batched-editing regime ($T\sim 50$ in our experiments, $r_e r_{\mathrm{in}}$ in the thousands), and $(r_e r_{\mathrm{in}})\times(r_e r_{\mathrm{in}})$ when $T$ is large. The contrast with~\S\ref{subsec:woodbury-memit} is sharp: the unstructured Woodbury solve carried per-edit design vectors of dimension $E\,d_{\mathrm{hidden}}$, and the global one-shot solve required factorizing a system of size $E\,d_{\mathrm{hidden}}\times E\,d_{\mathrm{hidden}}$. The Tucker reparameterization shrinks both the optimization variable and the per-edit witness by orders of magnitude, while the Woodbury identity preserves the small inversion size in batched mode.

\paragraph{Writeback.}
Given $G_{\mathrm{flat}}^{\star}$, the per-expert raw-space update is reconstructed via~\eqref{eq:per-expert-update} and composed with the per-expert null-space projector to yield the actual weight increment
\begin{equation}
\label{eq:writeback}
W_j \;\leftarrow\; W_j \;+\; U_{\mathrm{out}}\Bigl(\sum_{a=1}^{r_e} U_e[j,a]\,G_a^{\star}\Bigr) U_{\mathrm{in}}^{\!\top}\,P_j.
\end{equation}
For backbones whose down-projection is stored in transposed packed layout (e.g.\ Qwen3's \texttt{down\_proj} of shape $[E,d_{\mathrm{hidden}},d_{\mathrm{model}}]$), the increment in~\eqref{eq:writeback} is transposed before assignment; for \texttt{ModuleList}-style experts, the assignment is direct into \texttt{experts[j].down\_proj.weight}. Additional ablation on presence of null-space projection in Appendix \ref{appendix:null-space} shows it's necessity.

\paragraph{Spreading updates.} Iterating~\eqref{eq:writeback} across the critical layers $\ell\in\{L_0,\dots,L\}$ with the MEMIT-style residual schedule of~\eqref{eq:memit-spread} leads to calculating $G^{\star,L}$ only at the last layer $L$ and spreading it across previous layers from editing set with the same schedule:

\begin{equation}
\label{eq:our-spread}
G^{\star,\ell} \;=\; \frac{G^{\star,L}}{L-\ell+1},
\qquad
\hat{\Delta}^{\ell,\star} \;=\; U_{\mathrm{out}}^\ell\Bigl(\sum_{a=1}^{r_e} U^\ell_e[j,a]\,G_a^{\star, \ell}\Bigr) (U_{\mathrm{in}}^{\ell})^\top\,P^l_j,
\end{equation}

Doing so, factors $U^\ell_\mathrm{in}, U^\ell_\mathrm{out}, U^\ell_\mathrm{e}$ return weight perturbations into each layer's space and provide correct update, while saving time up to $n$ times, where $n$ is a number of edited layers.

\section{Experiments}\label{sec:experiments}

\begin{table*}[h!]
\centering
\small
\setlength{\tabcolsep}{6pt}
\resizebox{1.0\linewidth}{!}{
\begin{tabular}{clcccccccc}
\toprule
\multirow{2}{*}{Model} & \multirow{2}{*}{Method}
& \multicolumn{4}{c}{COUNTERFACT} 
& \multicolumn{4}{c}{ZsRE} \\
\cmidrule(lr){3-6} \cmidrule(lr){7-10}
& & Eff.$\uparrow$ & Gen.$\uparrow$ & Spe.$\uparrow$ & Uti.$\uparrow$
& Eff.$\uparrow$ & Gen.$\uparrow$ & Spe.$\uparrow$ & Uti.$\uparrow$ \\
\midrule
\multirow{7}{*}{\rotatebox{90}{Qwen3-30B-A3B}} & 
Pre-edited 
& 13.30\err{0.34} & 15.10\err{0.31} & 84.45\err{0.24} & 37.62
& 41.30\err{0.29} & 40.50\err{0.28} & 40.91\err{0.27} & 40.90 \\

& FT 
& 80.70\err{0.39} & 63.95\err{0.43} & 41.44\err{0.39} & 62.03
& 6.44\err{0.14} & 6.13\err{0.14} & 2.15\err{0.06} & 4.91 \\

& FT-L 
& 82.40\err{0.38} & 22.75\err{0.33} & 71.48\err{0.25} & 58.88
& 44.19\err{0.29} & 42.46\err{0.29} & 41.92\err{0.27} & 42.86 \\

& AdaLoRA 
& 51.90\err{0.50} & 49.75\err{0.40} & 48.10\err{0.26} & 49.92
& 3.66\err{0.09} & 3.60\err{0.09} & 4.68\err{0.10} & 3.98 \\

& UnKE 
& 89.30\err{0.31} & \underline{82.85\err{0.33}} & 48.15\err{0.33} & 73.43
& 31.43\err{0.28} & 29.78\err{0.27} & 25.30\err{0.23} & 28.84 \\

& MoEEdit 
& \textbf{97.90\err{0.14}} & \textbf{86.50\err{0.30}} & \textbf{83.45\err{0.23}} & \textbf{89.28}
& \textbf{75.29\err{0.28}} & \textbf{67.26\err{0.32}} & \textbf{43.18\err{0.27}} & \textbf{61.91} \\

& MoTE
& \underline{97.00\err{0.17}} & 82.60\err{0.33} & \underline{79.57\err{0.26}} & \underline{86.39}
& \underline{72.60\err{0.28}} & \underline{63.93\err{0.32}} & \underline{42.67\err{0.28}} & \underline{59.73} \\
\midrule

\multirow{7}{*}{\rotatebox{90}{GPT-OSS-20B}} & 
Pre-edited 
& 11.80\err{0.32} & 14.70\err{0.31} & 84.53\err{0.24} & 37.01
& 33.20\err{0.28} & 32.14\err{0.28} & 28.02\err{0.00} & 31.12 \\

& FT 
& 83.40\err{0.37} & \textbf{58.40\err{0.42}} & 55.72\err{0.33} & 65.84
& 25.57\err{0.28} & 23.41\err{0.26} & 17.61\err{0.21} & 22.20 \\

& FT-L 
& 73.80\err{0.44} & 43.10\err{0.48} & 59.75\err{0.33} & 58.88
& 32.75\err{0.29} & 33.09\err{0.30} & 30.06\err{0.26} & 31.97 \\

& AdaLoRA 
& 62.40\err{0.48} & \underline{55.00\err{0.42}} & 43.65\err{0.34} & 53.68
& 43.46\err{0.30} & 42.96\err{0.30} & \underline{33.60}\err{0.24} & 40.01 \\

& UnKE 
& 78.00\err{0.41} & 44.40\err{0.42} & 73.91\err{0.28} & 65.44
& 46.58\err{0.31} & 43.99\err{0.31} & 31.40\err{0.26} & 40.66 \\

& MoEEdit 
& \textbf{96.90\err{0.17}} & 38.80\err{0.41} & \underline{80.93\err{0.26}} & \textbf{72.21}
& \textbf{68.42\err{0.33}} & \underline{59.15\err{0.35}} & 33.23\err{0.27} & \underline{53.60} \\

& MoTE
& \underline{94.60\err{0.22}} & 37.55\err{0.41} & \textbf{81.85\err{0.25}} & \underline{71.33}
& \underline{66.42\err{0.34}} & \textbf{60.03\err{0.35}} & \textbf{37.11\err{0.28}} & \textbf{54.52} \\
\midrule

\multirow{7}{*}{\rotatebox{90}{Qwen3.6-35B-A3B}} & 
Pre-edited 
& 15.10\err{0.33} & 16.42\err{0.30} & 85.93\err{0.23} & 39.15
& 44.28\err{0.28} & 43.61\err{0.27} & 42.87\err{0.26} & 43.59 \\

& FT 
& 84.75\err{0.36} & 66.30\err{0.41} & 44.92\err{0.37} & 65.32
& 8.15\err{0.15} & 7.42\err{0.14} & 3.84\err{0.08} & 6.47 \\

& FT-L 
& 85.93\err{0.35} & 26.81\err{0.35} & 73.62\err{0.24} & 62.12
& 47.88\err{0.28} & 45.90\err{0.28} & 37.56\err{0.26} & 43.78 \\

& AdaLoRA 
& 56.42\err{0.47} & 52.91\err{0.39} & 50.84\err{0.25} & 53.39
& 7.43\err{0.12} & 7.05\err{0.11} & 6.88\err{0.11} & 7.12 \\

& UnKE 
& 87.84\err{0.29} & \underline{74.77\err{0.31}} & 52.10\err{0.31} & 71.57
& 35.82\err{0.27} & 33.74\err{0.26} & 28.91\err{0.22} & 32.82 \\

& MoEEdit 
& \textbf{93.40\err{0.25}} & \textbf{78.55\err{0.36}} & \textbf{77.99\err{0.26}} & \textbf{83.31}
& \textbf{61.39\err{0.32}} & \textbf{54.89\err{0.33}} & \textbf{40.93\err{0.27}} & \textbf{52.40} \\

& MoTE  
& \underline{89.20\err{0.31}} & 68.44\err{0.39} & \underline{75.33\err{0.27}} & \underline{77.66}
& \underline{58.38\err{0.43}} & \underline{51.79\err{0.22}} & \underline{39.03\err{0.32}} & \underline{49.73} \\
\bottomrule
\end{tabular}
}
\caption{Results on COUNTERFACT and ZsRE. The best results are in \textbf{bold}, the second best are \underline{underlined}}
\label{table:main}
\end{table*}

\subsection{Setup}

\textbf{Models}. Following the MoEEdit \cite{gu2026moeedit}, we evaluate three modern MoE LLMs: \textbf{Qwen3-30B-A3B-2507} \cite{qwen3technicalreport} (contains 128 experts; activates top-8 per token), \textbf{GPT-OSS-20B} \cite{openai2025gptoss120bgptoss20bmodel} (contains 32 experts; activates top-4 per token), and additionally \textbf{Qwen3.6-35B-A3B} \cite{qwen36_35b_a3b} (contains 256 experts; activates top-8 per token + 1 shared expert). Hyperparameters of our method for each model can be found in Appendix \ref{appendix:hyperparameters}.

\textbf{Datasets.} We evaluate performance on two common knowledge editing benchmarks: \textbf{COUNTERFACT} (single-hop counterfactual edit dataset) presented with MEMIT \cite{meng2022mass} and \textbf{ZsRE} (zero-shot relation extraction) \cite{levy2017zero}.

\textbf{Baselines.} We compare our method mostly with MoEEdit framework \cite{gu2026moeedit} since it is the only known method specifically targeting editing of MoE models. As additional competitors, we utilize simple fine-tuning (\textbf{FT}), fine-tuning with $L_\infty$-norm constraints (\textbf{FT-L})\cite{zhu2020modifying}, AdaLoRA \cite{zhang2023adalora}, and UnKE \cite{deng2024everything}.

\textbf{Metrics.} Reported metrics are standard for the field \cite{meng2022locating, meng2022mass}. They include (i) \textbf{Efficacy} (edit success on edited prompts), (ii) \textbf{Generalization} (success on paraphrased prompts), and (iii) Specificity (locality measured as a success on an unrelated, neighborhood prompts). To measure overall balance between them, we additionally report \textbf{Utility} (mean of three main metrics).

\subsection{Main results}

Main metrics for each model and method are shown in Table \ref{table:main}. Results show that our method provides second best edit quality in all setups and even outperform MoEEdit in some scenarios (e.g. GPT-OSS-20B model and ZsRE dataset). With GPT-OSS-20B model our method show higher locality on both datasets, which means better preserving of original model performance. Loss of 0.9-4.2 points by our method is highly acceptable due to the extreme speedup, which is shown in Table \ref{table:solver_layers_ablations}.

\subsection{Tucker vs no Tucker}
In this section we try to directly adapt MEMIT-like multilayer spreading to BCD and Global solution of \eqref{eq:moe-edit-obj} as shown in \eqref{eq:memit-spread}. Results in Table \ref{table:solver_layers_ablations} show that such naive approach fails to perform on par MoEEdit and MoTE. More detailed discussion can be found in Section\ref{subsec:woodbury-memit}.

\label{sec:exp tucker}
\begin{table*}[h!]
\centering
\small
\setlength{\tabcolsep}{6pt}
\begin{tabular}{clccccc}
\toprule
\multirow{2}{*}{Model} & \multirow{2}{*}{Method}
& \multicolumn{4}{c}{COUNTERFACT} & \multirow{2}{*}{Time,s $\downarrow$}  \\
\cmidrule(lr){3-6}
& & Eff.$\uparrow$ & Gen.$\uparrow$ & Spe.$\uparrow$ & Uti.$\uparrow$ \\
\midrule
\multirow{4}{*}{\rotatebox{0}{Qwen3-30B-A3B}} & 
MoEEdit 
& \textbf{97.90\err{0.14}} & \textbf{86.50\err{0.30}} & 83.45\err{0.23} & \textbf{89.28} & 477.0\err{17.7}\\

& BCD + \emph{speedup}
& 83.53\err{0.33} & 69.80\err{0.40} & \underline{83.58\err{0.23}} & 78.97 & 104.6\err{25.2}\\

& Global + \emph{speedup}
& 94.80\err{0.22} & 69.65\err{0.40} & \textbf{83.70\err{0.23}} & 82.72 & \underline{81.6\err{2.6}} \\

& MoTE
& \underline{97.00\err{0.17}} & \underline{82.60\err{0.33}} & 79.57\err{0.26} & \underline{86.39} & \textbf{76.8\err{7.7}} \\
\midrule

\multirow{4}{*}{\rotatebox{0}{GPT-OSS-20B}} & 
MoEEdit 
& \textbf{96.90\err{0.17}} & 38.80\err{0.41} & 80.93\err{0.26} & \textbf{72.21} & 469.2\err{28.0}\\

& BCD + \emph{speedup}
& 89.5\err{0.19} & \textbf{40.85\err{0.42}} & 81.73\err{0.25} & 70.69 & 123.9\err{2.0} \\

& Global + \emph{speedup}
& 89.3\err{0.22} & \underline{40.55\err{0.43}} & \underline{81.74\err{0.28}} & 70.53 & \underline{100.8\err{2.2}}\\

& MoTE
& \underline{94.60\err{0.22}} & 37.55\err{0.41} & \textbf{81.85\err{0.25}} & \underline{71.33} & \textbf{88.9\err{1.3}}\\
\midrule

\multirow{4}{*}{\rotatebox{0}{Qwen3.6-35B-A3B}} &
MoEEdit 
& \textbf{93.40\err{0.25}} & \textbf{78.55\err{0.36}} & 77.79\err{0.26} & \textbf{83.25} &  717.4\err{23.8}\\

& BCD + \emph{speedup}
& 82.40\err{0.38} & 53.75\err{0.43} & \textbf{81.40\err{0.24}} & 72.52 & 208.0\err{4.4}\\

& Global + \emph{speedup}
& 82.30\err{0.38} & 54.35\err{0.43} & \underline{81.33\err{0.24}} & 72.66 & \underline{121.0\err{4.5}}\\

& MoTE
& \underline{89.20\err{0.31}} & \underline{68.40\err{0.40}} & 75.33\err{0.27} & \underline{77.64} & \textbf{112.5\err{1.3}} \\
\bottomrule
\end{tabular}
\caption{Results on editing 1000 facts from COUNTERFACT dataset with different solvers and multi-layer modes. Time is measured when all $v^*$ vectors are already computed. The best results are in \textbf{bold}, the second best are \underline{underlined}. \emph{Speedup} stands for spreding updates as in ~\eqref{eq:our-spread} instead of spreading residuals as in ~\eqref{eq:memit-spread}.}
\label{table:solver_layers_ablations}
\end{table*}

\section{Limitations and discussion}

Knowledge editing methods can be used both for lightweight model updates and for poisoning models forcing them into malicios behavior. It naturally arises ethical concerns about usage of such algorithms, so we call for their responsible use.

Limitations of our method MoTE include (i) high computational cost with large batch sizes ($>>1000$), (ii) need of computing HOSVD for making Tucker Decomposition.

\section{Conclusion}
\label{sec:conclusion}
Beyond the immediate practical benefit of a fast, principled editor for MoE LLMs, our results carry a broader message: closed-form, parameter-modifying knowledge editing is not intrinsically tied to dense architectures. With the right algebraic treatment of the layer's structure, the same family of techniques can be extended to substantially more complex architectures than the dense FFNs in which they were originally developed. We view this as an encouraging sign that the principled, optimization-based branch of the KE literature can keep pace with the architectural evolution of modern LLMs.

\end{mainpart}

\begin{appendixpart}

\section{Metrics measurement}\label{appendix:metrics}

\subsection{ZsRE Evaluation Protocol}

Following prior work on model editing, we define a language model $\mathcal{M}$, an edit request $(s_i, r_i)$, a desired target response $o_\text{new}$, and the model's original prediction $o_\text{true}$.

\paragraph{Efficacy.}
Edit success measures whether the edited model greedy predicts the target answer on the modified prompt:
\begin{equation*}
\mathbb{E}_i
\left[
\mathbf{1}
\left(
o_\text{new} =
\arg\max_{t}
P_{\mathcal{M}}
(t \mid (s_i, r_i))
\right)
\right].
\end{equation*}

\paragraph{Generalization.}
We evaluate whether the edit transfers to paraphrased variants
$\widehat{\mathcal{P}}(s_i, r_i)$ of the original prompt:
\begin{equation*}
\mathbb{E}_i
\left[
\mathbf{1}
\left(
o_\text{new} =
\arg\max_{t}
P_{\mathcal{M}}
(t \mid \widehat{\mathcal{P}}(s_i, r_i))
\right)
\right].
\end{equation*}

\paragraph{Specificity.}
Locality preservation quantifies whether unrelated prompts
$\mathcal{U}(s_i, r_i)$ remain unaffected after editing:
\begin{equation*}
\mathbb{E}_i
\left[
\mathbf{1}
\left(
o_\text{true} =
\arg\max_{t}
P_{\mathcal{M}}
(t \mid \mathcal{U}(s_i, r_i))
\right)
\right].
\end{equation*}

\subsection{CounterFact Evaluation Protocol}

For CounterFact, we adopt probability-based metrics under the same setup
$(s_i, r_i)$ with target answer $o_\text{new}$ and original prediction $o_\text{true}$.

\paragraph{Efficacy.}
This metric evaluates whether the edited model assigns higher likelihood
to the target answer than to the original prediction:
\begin{equation*}
\mathbb{E}_i
\left[
P_{\mathcal{M}}
(o_\text{new} \mid (s_i, r_i))
>
P_{\mathcal{M}}
(o_\text{true} \mid (s_i, r_i))
\right].
\end{equation*}

\paragraph{Generalization.}
We measure whether paraphrased prompts
$\widehat{\mathcal{P}}(s_i, r_i)$ preserve the edited knowledge:
\begin{equation*}
\mathbb{E}_i
\left[
P_{\mathcal{M}}
(o_\text{new} \mid \widehat{\mathcal{P}}(s_i, r_i))
>
P_{\mathcal{M}}
(o_\text{true} \mid \widehat{\mathcal{P}}(s_i, r_i))
\right].
\end{equation*}

\paragraph{Specificity.}
For neighboring prompts $\mathcal{U}(s_i, r_i)$ involving related but distinct entities,
specificity measures whether the original behavior is preserved:
\begin{equation*}
\mathbb{E}_i
\left[
P_{\mathcal{M}}
(o_\text{true} \mid \mathcal{U}(s_i, r_i))
>
P_{\mathcal{M}}
(o_\text{new} \mid \mathcal{U}(s_i, r_i))
\right].
\end{equation*}

\section{Methods hyperparameters}\label{appendix:hyperparameters}
\paragraph{Hardware and Precision.}
All experiments are executed on a single machine equipped with an NVIDIA H200 GPU.
Model parameters are stored in BF16 precision to reduce memory usage, while all optimizer
states, collected activations and gradient updates are maintained in FP32 for improved numerical stability.

\paragraph{Fine-Tuning Baselines.}
We evaluate both standard fine-tuning (FT) and constrained fine-tuning (FT-L).
The FT-L variant restricts the magnitude of parameter updates through a $L_\infty$-norm constraint $\varepsilon$.

For all three models Qwen3-30B-A3B, GPT-OSS-20B and Qwen3.6-35B-A3B we use
$\varepsilon = 10^{-3}$ and a learning rate of $10^{-3}$.
Parameter updates are applied to last layer presented in Table \ref{table:hyperparameters}.

Training is conducted for 25 epochs with both weight decay and KL regularization disabled.

\paragraph{UnKE.}
Since UnKE relies on a two-stage optimization procedure,
we configure separate hyperparameters for each phase.

For both Qwen models, the first stage uses a learning rate of
$5 \times 10^{-1}$ with 25 optimization steps and weight decay
$10^{-3}$.
The second stage adopts a learning rate of $2 \times 10^{-4}$
for 50 additional steps.

For GPT-OSS-20B, the first stage also employs a learning rate of
$5 \times 10^{-1}$ but increases the number of optimization steps to 50,
while retaining the same weight decay coefficient.
The second stage uses a learning rate of $10^{-4}$ with 50 optimization steps.

For consistency, all UnKE updates are restricted to the same last layer, as the other baselines.
Optimization is performed on the final subject token, following the original structured editing setup.

\paragraph{AdaLoRA.}
For AdaLoRA, trainable adapters are inserted into all transformer layers.
We set the scaling factor $\alpha = 32$ and rank $r = 8$.

The learning rate is set to $5 \times 10^{-3}$ for Qwen models
and $5 \times 10^{-4}$ for GPT-OSS-20B.
Optimization is performed for 25 update steps for all models.

\paragraph{MoEEdit.}
Layers modified with this method are the same as we show in Table \ref{table:hyperparameters}.

On Qwen3-30B-A3B, optimization is performed for 25 steps with learning rate $0.1$ and 4 Block Coordinate Descent (BCD) iterations. For Qwen3.6-35B-A3B we reduce number of iterations by half due to twice as large number of experts.
For GPT-OSS-20B, we use 50 optimization steps, learning rate $0.2$, and 10 BCD iterations.

For all models, we set the regularization coefficient
$\lambda = 1$ and the KL penalty weight to $0.0625$.

To estimate the covariance matrix used in null-space projection,
we sample 100K activations.
The projection threshold is fixed to $0.02$.

\paragraph{MoTE.}
For this method we utilize the same target vector$v^*$ as for MoEEdit, so all related parameters remain the same. Main differences are shown in Table \ref{table:hyperparameters}.
\begin{table*}[h!]
    \centering
    \small
    \begin{tabular}{cccc}
    \toprule
         Model & $\lambda$ & Layers & $\varepsilon_\text{whitening}$ \\
         \midrule
         Qwen3-30B-A3B & 0.1 & {3, 4, 5, 6, 7} & $1 \times10^{-5}$ \\
         GPT-OSS-20B & 1 & {3, 4, 5} & $1 \times10^{-2}$ \\
         Qwen3.6-35B-A3B & 0.01 & {3, 4, 5, 6} & $1 \times10^{-5}$ \\
    \bottomrule
    \end{tabular}
    \caption{Hyperparameters for MoTE method}
    \label{table:hyperparameters}
\end{table*}

\section{Additional Experiments}
\subsection{Routing shift results}
Following \cite{gu2026moeedit}, we conduct routing shift distribution analysis on each model and \textbf{COUNTERFACT} dataset. We perform 1000 edits with batch size 50 and measure routing similarity (RS) on both editing set and preserved 1000 edits set, averaging score over windows of 10 layers. Results in Table \ref{table:rs} show that our method outperforms all methods except MoEEdit. The most probable reason is Tucker Decomposition itself, because it projects inputs and outputs into core space, which shifts the distribution by design.
\begin{table}[h!]
\centering
\small
\begin{tabular}{clcccccc}
\toprule
\multirow{2}{*}{\textbf{Model}} & \multirow{2}{*}{\textbf{Method}} & \multicolumn{3}{c}{\textbf{Editing Set RS}$\uparrow$} & \multicolumn{3}{c}{\textbf{Preservation Set RS}$\uparrow$} \\
\cmidrule(lr){3-5} \cmidrule(lr){6-8}
& & Early & Middle & Late & Early & Middle & Late \\
\midrule
\multirow{6}{*}{\rotatebox{0}{Qwen3-30B-A3B}} & FT & 23.57 & 26.58 & 29.98 & 24.72 & 27.45 & 30.97 \\
                                                & FT-L & 47.01 & 51.20 & 53.68 & 48.80 & 50.17 & 53.45 \\
                                                & AdaLoRA & 16.63 & 24.11 & 27.00 & 16.38 & 23.84 & 26.60 \\
                                                & UnKE & 52.46 & 44.12 & 44.80 & 49.90 & 41.91 & 43.84 \\
                                                & MoEEdit & \textbf{86.62} & \textbf{88.16} & \textbf{89.93} & \textbf{87.02} & \textbf{88.55} & \textbf{90.22} \\
                                                & Ours  & \underline{59.82} & \underline{69.11} & \underline{73.88} & \underline{67.41} & \underline{69.83} & \underline{66.46} \\
\midrule
\multirow{6}{*}{\rotatebox{0}{GPT-OSS-20B}} & FT & 31.75 & 28.09 & 35.25 & 32.94 & 28.10 & 34.98 \\
                                                & FT-L & 60.51 & 57.01 & 64.95 & 61.30 & 57.11 & 64.12 \\
                                                & AdaLoRA & 22.94 & 18.00 & 25.91 & 22.11 & 19.06 & 25.28 \\
                                                & UnKE & 57.15 & 53.64 & 60.86 & 57.77 & 53.15 & 60.26 \\
                                                & MoEEdit & \textbf{77.37} & \textbf{75.56} & \textbf{80.83} & \textbf{78.00} & \textbf{76.13} & \textbf{80.53} \\
                                                & Ours  & \underline{72.06} & \underline{68.95} & \underline{75.01} & \underline{72.38} & \underline{68.78} & \underline{74.83} \\
\midrule
\multirow{6}{*}{\rotatebox{0}{Qwen3.6-35B-A3B}} & FT & 33.83 & 30.27 & 37.39 & 34.45 & 30.94 & 36.98 \\
                                                & FT-L & 62.92 & 59.15 & 66.63 & 63.48 & 59.80 & 66.12 \\
                                                & AdaLoRA & 24.08 & 20.49 & 27.77 & 24.61 & 21.06 & 27.28 \\
                                                & UnKE & 59.34 & 55.02 & 62.81 & 59.89 & 55.55 & 62.26 \\
                                                & MoEEdit & \textbf{79.00} & \textbf{77.20} & \textbf{82.50} & \textbf{79.50} & \textbf{77.55} & \textbf{82.05} \\
                                                & Ours  & \underline{73.52} & \underline{70.45} & \underline{76.55} & \underline{74.02} & \underline{70.48} & \underline{76.03} \\
\bottomrule
\end{tabular}
\caption{Routing similarity before and after edit on editing and preservation sets. For Qwen model Early Middle and Late sites are layer 11-20, 21-30 and 31-40, for GPT-OSS: 5-9, 10-14 and 15-19, respectively. Best results are in \textbf{bold}, second best are \underline{underlined}}
\label{table:rs}
\end{table}

\subsection{Whitening ablation} \label{appendix:whitening}
Following results of \cite{yuebintd}, we made an ablation on whitening type for each model. As for other experiments, we make 1000 edits from \textbf{COUNTERFACT} dataset with batch size 50 and other hyperparameters exactly matching those from Appendix \ref{appendix:hyperparameters}. Results in Table \ref{table:whitening} show that using only \textit{in} whitening show best performance and it's absence lead to huge model degradation. \textit{Out} whitening in original implementation with random gradients doesn't yield any improvement and even brings slight decline in main metrics in combination with \textit{in} whitening.

\begin{table*}[h!]
\centering
\small
\setlength{\tabcolsep}{6pt}
\begin{tabular}{cccccc}
\toprule
\multirow{2}{*}{Model} & \multirow{2}{*}{Whitening type}
& \multicolumn{4}{c}{COUNTERFACT} \\
\cmidrule(lr){3-6}
& & Eff.$\uparrow$ & Gen.$\uparrow$ & Spe.$\uparrow$ & Uti.$\uparrow$ \\
\midrule
\multirow{4}{*}{\rotatebox{0}{Qwen3-30B-A3B}} & 
None 
& 63.50\err{0.48} & 61.15\err{0.42} & 66.45\err{0.31} & 63.70 \\

& Out
& 66.50\err{0.47} & 61.95\err{0.42} & \textbf{86.06}\err{0.31} & 71.50 \\

& In
& \textbf{97.00}\err{0.17} & \textbf{82.60}\err{0.33} & \underline{79.57}\err{0.26} & \textbf{86.39} \\

& Both
& \underline{96.80}\err{0.18} & \underline{82.55}\err{0.32} & 79.47\err{0.26} & \underline{86.27}  \\

\midrule
\multirow{4}{*}{\rotatebox{0}{GPT-OSS-20B}} & 
None 
& 41.70\err{0.49} & 37.10\err{0.42} & 65.50\err{0.33} & 48.10 \\

& Out
& 45.87\err{0.50} & 36.80\err{0.41} & 66.32\err{0.33} & 49.66 \\

& In
& \textbf{94.60}\err{0.22} & \underline{37.55}\err{0.41} & \textbf{81.85}\err{0.25} & \underline{71.33} \\

& Both
& \underline{94.50}\err{0.23} & \textbf{38.30}\err{0.42} & \underline{81.76}\err{0.25} & \textbf{71.52}  \\

\midrule
\multirow{4}{*}{\rotatebox{0}{Qwen3.6-35B-A3B}} &
None 
& \underline{82.50}\err{0.38} & \underline{73.00}\err{0.37} & 64.50\err{0.32} & 73.33 \\

& Out
& 82.10\err{0.38} & \textbf{73.95}\err{0.37} & 65.07\err{0.30} & 73.71 \\

& In
& \textbf{89.20}\err{0.31} & 68.44\err{0.39} & \underline{75.33}\err{0.27} & \textbf{77.66} \\

& Both
& 88.50\err{0.32} & 68.00\err{0.40} & \textbf{75.52}\err{0.27} & \underline{77.34}  \\

\bottomrule
\end{tabular}
\caption{Results of editing 1000 facts from \textbf{COUNTERFACT} dataset with different whitening types. Best results are in \textbf{bold}, second best are \underline{underlined}}
\label{table:whitening}
\end{table*}

\subsection{Null-space ablation} \label{appendix:null-space}
In this section we compare performance of MoTE method with enabled and disabled null-space projection. From results shown in Table \ref{table:nullspace} and evaluation protocol of \textbf{COUNTERFACT} \ref{appendix:metrics}, we can conclude, that without null-space projection our method turn model into a random predictor which chooses between $o_\text{true}$ and $o_\text{new}$. It shows the urgency of this component for preserving model overall performance.
\begin{table*}[h!]
\centering
\small
\setlength{\tabcolsep}{6pt}
\begin{tabular}{cccccc}
\toprule
\multirow{2}{*}{Model} & \multirow{2}{*}{Null-space}
& \multicolumn{4}{c}{COUNTERFACT} \\
\cmidrule(lr){3-6}
& & Eff.$\uparrow$ & Gen.$\uparrow$ & Spe.$\uparrow$ & Uti.$\uparrow$ \\
\midrule
\multirow{2}{*}{\rotatebox{0}{Qwen3-30B-A3B}} & 

$\checkmark$
& 97.00\err{0.17} & 82.60\err{0.33} & 79.57\err{0.26} & 86.39 \\

& $\times$
& 49.60\err{0.50} & 49.0\err{0.36} & 48.91\err{0.21} & 49.17  \\

\midrule
\multirow{2}{*}{\rotatebox{0}{GPT-OSS-20B}} & 
$\checkmark$
& 94.60\err{0.22} & 37.55\err{0.41} & 81.85\err{0.25} & 71.33 \\

& $\times$
& 50.10\err{0.50} & 50.90\err{0.47} & 48.44\err{0.45} & 49.81 \\

\midrule
\multirow{2}{*}{\rotatebox{0}{Qwen3.6-35B-A3B}} &

$\checkmark$
& 89.20\err{0.31} & 68.44\err{0.39} & 75.33\err{0.27} & 77.66 \\

& $\times$
& 49.90\err{0.52} & 51.1\err{0.48} & 50.3\err{0.55} & 50.43 \\

\bottomrule
\end{tabular}
\caption{Results of editing 1000 facts from \textbf{COUNTERFACT} dataset with MoTE method with \textit{in} whitening and with or without null-space projection}
\label{table:nullspace}
\end{table*}

\end{appendixpart}

\end{document}